%% file: main.tex
\documentclass[letterpaper]{article} 
\usepackage{aaai25}  
\usepackage{times}  
\usepackage{helvet}  
\usepackage{courier}  
\usepackage[hyphens]{url}  
\usepackage{graphicx} 
\def\UrlFont{\rm}  
\usepackage{natbib}  
\usepackage{caption} 
\usepackage{algorithm}
\usepackage{algorithmic}
\usepackage{pifont}
\usepackage{multirow}
\usepackage{booktabs}
\usepackage{amssymb}
\usepackage{dsfont}

\usepackage{footmisc}
\DefineFNsymbols{mySymbols}{{\ensuremath\dagger}{\ensuremath\ddagger}\S\P
   *{**}{\ensuremath{\dagger\dagger}}{\ensuremath{\ddagger\ddagger}}}
\setfnsymbol{mySymbols}

\usepackage{newfloat}
\usepackage{listings}
\DeclareCaptionStyle{ruled}{labelfont=normalfont,labelsep=colon,strut=off} 
\floatstyle{ruled}
\newfloat{listing}{tb}{lst}{}
\floatname{listing}{Listing}
\usepackage{amsmath,amsfonts,bm}
\usepackage{xcolor}
\definecolor{brickred}{rgb}{0.8, 0.25, 0.33}
\definecolor{brickgreen}{rgb}{0.25, 0.8, 0.33}
\newcommand{\cm}{\textcolor{brickred}{\ding{51}}}%
\newcommand{\xm}{\textcolor{brickgreen}{\ding{55}}}

\def\eqref#1{equation~\ref{#1}}

\def\1{\bm{1}}

\def\vtheta{{\bm{\theta}}}

\def\vm{{\bm{m}}}

\def\vx{{\bm{x}}}

\def\vz{{\bm{z}}}

\DeclareMathAlphabet{\mathsfit}{\encodingdefault}{\sfdefault}{m}{sl}
\SetMathAlphabet{\mathsfit}{bold}{\encodingdefault}{\sfdefault}{bx}{n}

\def\gA{{\mathcal{A}}}

\def\gD{{\mathcal{D}}}

\def\gM{{\mathcal{M}}}
\def\gN{{\mathcal{N}}}

\title{SeeDiff: Off-the-Shelf Seeded Mask Generation from Diffusion Models}
\author{
    Joon Hyun Park\textsuperscript{\rm 1},
    Kumju Jo\textsuperscript{\rm 1},
    Sungyong Baik\textsuperscript{\rm 1, \rm 2}\thanks{Corresponding author.}
}
\affiliations{
     \textsuperscript{\rm 1} Dept. of Artificial Intelligence, Hanyang University, South Korea\\ \textsuperscript{\rm 2} Dept. of Data Science, Hanyang University, South Korea\\
    \{pjay258, juice0630, dsybaik\}@hanyang.ac.kr\\

}

\usepackage{bibentry}
\usepackage{dsfont}
\begin{document}

\maketitle

\begin{abstract}
Entrusted with the goal of pixel-level object classification, the semantic segmentation networks entail the laborious preparation of pixel-level annotation masks.
To obtain pixel-level annotation masks for a given class without human efforts, recent few works have proposed to generate pairs of images and annotation masks by employing image and text relationships modeled by text-to-image generative models, especially Stable Diffusion.
However, these works do not fully exploit the capability of text-guided Diffusion models and thus require a pre-trained segmentation network, careful text prompt tuning, or the training of a segmentation network to generate final annotation masks.
In this work, we take a closer look at attention mechanisms of Stable Diffusion, from which we draw connections with classical seeded segmentation approaches.
In particular, we show that cross-attention alone provides very coarse object localization, which however can provide initial seeds.
Then, akin to region expansion in seeded segmentation, we utilize the semantic-correspondence-modeling capability of self-attention to iteratively spread the attention to the whole class from the seeds using multi-scale self-attention maps.
We also observe that a simple-text-guided synthetic image often has a uniform background, which is easier to find correspondences, compared to complex-structured objects.
Thus, we further refine a mask using a more accurate background mask.
Our proposed method, dubbed \textbf{SeeDiff}, generates high-quality masks off-the-shelf from Stable Diffusion, without additional training procedure, prompt tuning, or a pre-trained segmentation network. 
\end{abstract}
\begin{links}  
\link{Code}{https://github.com/BAIKLAB/SeeDiff.git}
\end{links}

\section{Introduction}
As one of fundamental tasks in computer vision,  semantic segmentation has a broad range of applications from medical imaging~\cite{medicalimaging,medical2} to tracking~\cite{track-Anyting}.
The applicability and practicability of semantic segmentation have increased with the emergence of neural networks for segmentation~\cite{fcn,Unet}. 
However, the training of semantic segmentation networks is data-hungry in nature, entailing a large amount of fine-grained pixel-level annotation and even more so for segmentation foundation models (1.1 billion masks across 11 million images).
Such data-hunger nature can make the training of segmentation networks laborious, thereby limiting its practicability.

To reduce the annotation cost, several works have employed weakly supervised learning frameworks, where only coarse labels (e.g., bounding boxes~\cite{bbox}, lines or points~\cite{line}, or image-level class labels~\cite{image-level}, etc.) are available. 
However, relying on such weak labels has led to task-specific complex designs~\cite{seg2} and relatively low performance.

\begin{figure}[t]
\centering
\includegraphics[width=0.9\columnwidth]{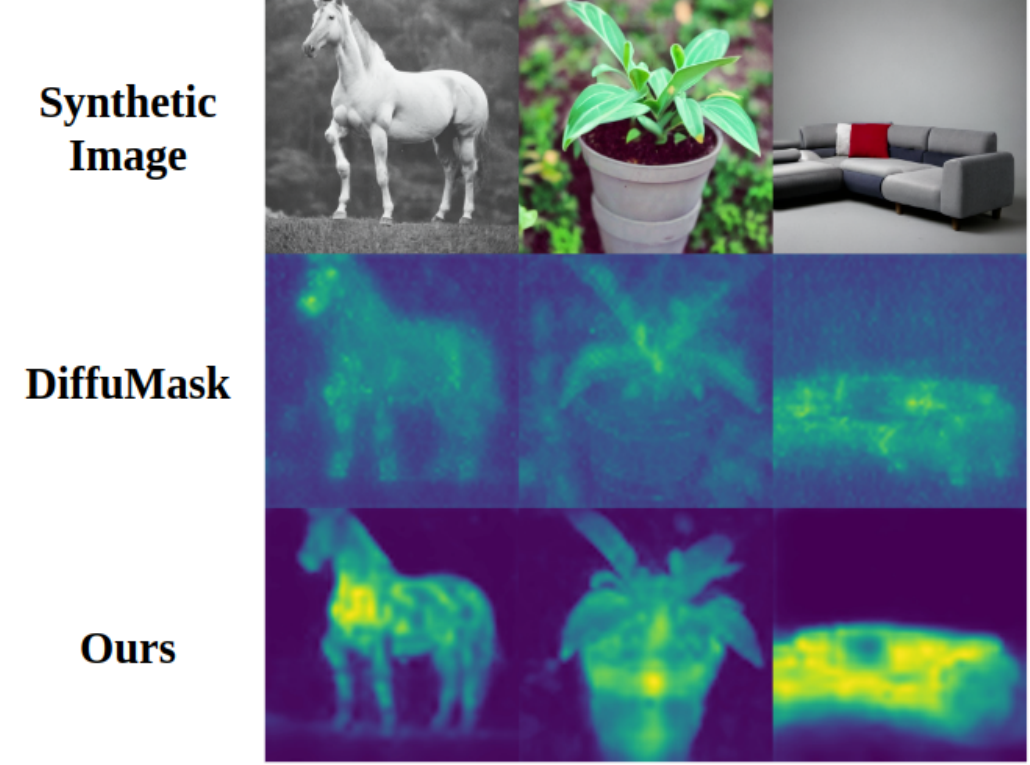} 
\caption{Comparisons against our baseline, DiffuMask~\cite{diffumask} on the quality of a generated mask.
 To better visualize the quality and accuracy of a generated mask, we visualize a soft mask before discretization.
We can see that our proposed method SeeDiff produces more accurate masks with sharp boundaries, compared to our baseline.
}
\label{fig:diff_sediff}
\end{figure}
Alternatively, there has been efforts on generating synthetic segmentation dataset consisting of image-mask pairs, in order to remove human efforts for annotation.
Early works~\cite{datasetgan,bigdatasetgan} have employed the features of generative adversarial networks (GANs) to generate synthetic data.
However, these methods introduce ad-hoc decoders that still require a few pixel-level annotations for training.

Following the advances in generative models~\cite{DDPM} with the emergence of diffusion models, few recent works~\cite{diffumask,Dataset-Diffusion} have shifted to using diffusion models to generate synthetic segmentation datasets.
They employ text-conditioned diffusion models (e.g., Stable Diffusion~\cite{LDM}), which can generate realistic images corresponding to input texts and thereby allowing them to easily control image generation.
In particular, previous works have mainly focused on image-text relationships learned by cross-attention layers.
Such multimodal latent space allows them to localize which image regions correspond to particular texts/classes, resulting in mask generation.
However, cross-attention maps, which the previous works heavily rely on, can provide very coarse masks.
These coarse masks result in prior works resorting to the utilization of a pre-trained segmentation network trained with pixel-level annotations~\cite{diffumask}, careful text prompt tuning~\cite{Dataset-Diffusion}, or extra training procedures~\cite{diffumask,Dataset-Diffusion}, which can result in poor quality masks for new classes that are unknown during the pre-training or additional training of segmentation networks and text prompt tuning.

In this work, we perform an analysis on attention maps produced by Stable Diffusion, in order to obtain high-quality masks.
Upon analysis, we claim that both cross-attention layers and self-attention layers play crucial yet complementary roles.
While cross-attention layers provide crude attention maps, cross-attention maps can serve as good starting points for localizing objects of interests corresponding to provided text.
On the other hand, self-attention layers are well-known for modeling correspondences, which can be used to find other parts of the object that cross-attention might have missed.
We also note different characteristics of self-attention across different layers: deeper layers provide low-resolution attention maps with better semantic information, while earlier layers give high-resolution fine-grained attention maps with less semantic information, corroborating previous findings~\cite{visualizing&understandingcnn,vit}.
We also observe that self-attention layers struggle to find correspondences between all parts of the whole object with non-homogeneous appearances, whereas self-attention layers find more accurate correspondences within background, which is usually uniform in a simple-text-guided synthetic image.

Founded upon our observations from analysis, we draw connections between the observed characteristics of attention layers and classical seeded segmentation algorithms~\cite{seeded_segmentation}.
Seeded segmentation first finds seeds (initial starting pixels) and expand segmentation masks to pixels that are similar to seeds.
With this realization, we propose a new off-the-shelf framework named \textbf{SeeDiff} that draws inspiration from seeded segmentation~\cite{seeded_segmentation} to utilize attention maps of Stable Diffusion to generate masks.
In particular, we localize a target class object that corresponds to a text, utilizing the cross-attention maps, which quantify text-image correspondences.
The coarse responses of the cross-attention maps provide initial seeds, which are then used to find other parts of the object.
Such mask region expansion is performed by using self-attention maps that correspond to seeds to find other missing parts of the object.
We perform mask region expansion iteratively, as the feature resolution increases, to exploit semantic information from low-resolution features and fine-grained details from high-resolution features.
We further refine a mask, using a background mask that can be obtained by mask region expansion from background seeds extracted from the inverted mask.

Experimental results demonstrate outstanding performance of segmentation networks trained with our generated image-mask pairs.
The results underline the effectiveness of our proposed approach in generating high-quality fine-grained pixel-level annotations, without the need for a pre-trained segmentation network, text prompt tuning, training a new module, or learning procedures.

\begin{figure*}[t]
\centering
\includegraphics[width=1.0\textwidth]{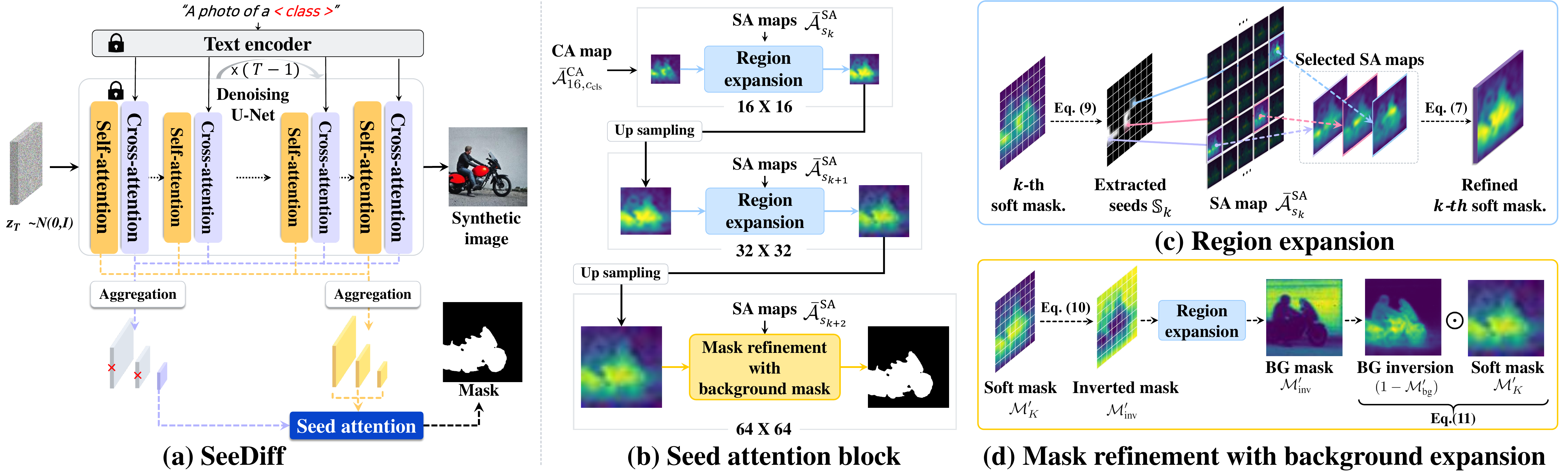} 
\caption{Overall framework: (a) Our method SeeDiff is an off-the-shelf mechanisms that generates a mask by utilizing extract attention maps from a pre-trained Stable Diffusion conditioned on an input prompt.
(b) Our mask generation draws inspiration from a classical seeded segmentation process, where a cross-attention (CA) map provides initial cues to the target object location, which are then iteratively expanded to object regions through (c) region expansion with SA maps.
(d) A mask is further refined with a background mask, which is similarly obtained via seed extraction and region expansion of an inverted mask.
}
\label{fig_overall}
\end{figure*}

\section{Related Works}

\textbf{Semantic Segmentation.}
Providing fine-grained yet essential information, semantic segmentation has gained attention not only from high-level computer vision tasks (e.g., scene understanding~\cite{sceneunderstanding}) but a broad range of other domains, such as self-driving cars~\cite{auto-driving} and medical image analysis~\cite{medicalimaging,medical2}.
Since the emergence of deep neural networks (DNNs), the performance has elevated with advances in neural network architecture designs~\cite{fcn,diatlednet,DeepLab,Unet,DeconvNet,SegNet,ParseNet,GCN}, especially with transformers~\cite{SegFormer,SETR,MaskFormer,Mask2Former} and foundation models~\cite{SAM,SEEM}.
However, behind the exceptional performance of DNN-based segmentation models lies a significant amount of human efforts in collecting and annotating large-scale data required for training DNNs.
This is especially the case for transformer-based networks and foundation models, which are known for its data hunger~\cite{SAM}.
To make things worse, segmentation networks require fine-grained pixel-level annotations, making dataset preparation for training even more laborious.
To reduce the annotation cost of semantic segmentation, several works have proposed to make use of cheaper annotations, such as image-wise category labels~\cite{bbox}, points/lines~\cite{line}, and bounding boxes~\cite{image-level}.
However, these weak labels lead to performance degradation.
As such, recent models, such as SAM and Mask2Former, still use pixel-level annotations to maximize performance.

\subsubsection{Dataset Generation for Semantic Segmentation.}
To minimize human efforts in collecting annotation, a line of works have tried to borrow the data synthesis capability of generative models to generate synthetic datasets consisting of image-mask pairs.
Early methods~\cite{LS-GAN,GAN-arxiv,datasetgan,bigdatasetgan} employ generative adversarial networks (GANs)~\cite{GAN}, where semantic features from intermediate layers are extracted and passed to need-to-train decoders for mask generation.
However, these works either require optimization process and can only perform foreground-background separation~\cite{LS-GAN,GAN-arxiv} or extra decoders~\cite{datasetgan,bigdatasetgan} that still need to be trained with few images annotated with pixel-level fine-grained masks.
Recent works~\cite{grounding,diffumask,datasetdm,Dataset-Diffusion} have shifted to diffusion-based~\cite{DDPM} text-to-image generative models (e.g., GLIDE~\cite{GLIDE}, Imagen~\cite{Imagen}, and Stable Diffusion~\cite{LDM}) for its breakthrough in generation performance and easy control over generation via text.
While using diffusion models has led to high-quality image-mask generation, they require either a pre-trained object detector~\cite{grounding} or a pre-trained segmentation network~\cite{diffumask}; extra modules that need to be trained~\cite{grounding,diffumask}; or extra learning processes~\cite{Dataset-Diffusion, DiffusionSeg} for refinement.
These processes can lead to poor quality mask generation for classes unknown during such processes, thereby being unable to fully exploit the capabilities of text-to-image generative models trained on a very large-scale dataset. 
Our work, on the other hand, draws inspiration from seeded segmentation, which we claim to be easily transferable to image-mask generation task with off-the-shelf text-guided diffusion models.
As a result, we synthesize high-quality image-mask datasets, without the need for pre-trained segmentation/detection networks, extra trainable modules, or optimization/learning processes.

\section{Background}
\subsection{Problem Formulation}
The objective of this work is to construct a high-quality synthetic segmentation dataset $\gD=\{\vx_n, \vm_n\}^{N}_{n=1}$ consisting of synthetic images $\vx_n$ with corresponding pixel-level masks $\vm_n$.
In order to have correspondences between generated images and masks, we first need to have control over which objects will be generated in images.
To this end, similar to previous works~\cite{diffumask,Dataset-Diffusion}, we generate images via a text-to-image generative model, which allows for a control over image generation via texts.
Following prior works~\cite{diffumask,Dataset-Diffusion}, we use instantiate a text-to-image generative model with Stable Diffusion~\cite{LDM}, which is one of recent, successful, and open-source text-to-image generative models.

\subsection{Stable Diffusion}
Stable Diffusion~\cite{LDM} is a text-guided diffusion model that aims to generate an image $\vx$ from a random noise $\vz\sim\gN(\mathbf{0},\mathbf{I})$, guided by a text embedding $\bm{\tau}(y)$ with an input text prompt $y$ of length $P$ and a text encoder $\bm{\tau}$.
Diffusion models assume that a random noise $\vz$ and $\vx$ are connected by Markov chain that gradually adds noise to $\vx=\vz_0$ to form $\vz=\vz_T$ over a pre-defined number of time steps $T$:
\begin{equation}
    q(\vz_t|\vz_{t-1}) = \gN(\vz_t;\sqrt{1-\beta_t}\vz_{t-1},\beta_t\mathbf{I}) \, ,
\end{equation}
where $\beta_t$ is a variance or noise schedule.
Under the assumption, diffusion models formulate image generation as a progressive noise removal (i.e., denoising) from a noisy image to obtain gradually less noisy image and eventually a clean image at the end.
Hence, a denoising network $\bm{\epsilon}_\vtheta$, parameterized by parameters $\vtheta$, is trained to estimate an added noise $\bm{\epsilon}\sim\gN(\mathbf{0}, \mathbf{I})$ guided by conditioning text prompt $y$ at each time step $t$, optimizing the following objective function:
\begin{equation}
    \mathbb{E}_{\vx,\bm{\epsilon}\sim\gN(\mathbf{0},\mathbf{I}),t,\bm{\tau}(y)}\left[\big\| \bm{\epsilon} - \bm{\epsilon}_\vtheta(\vz_t, t, \bm{\tau}(y)) \big\|^2_2 \right].
\end{equation}
Then, during image generation process, a noise estimated by a denoising network is removed from a noisy image to gradually obtain a less noisy image, leading to a new clean generated image after a pre-defined number of steps $T$.
In contrast to previous diffusion models, Stable Diffusion performs noise addition and denoising process on the latent space, instead of pixel space.

A denoising network in Stable Diffusion employs U-Net~\cite{Unet} architecture consisting of residual blocks and transformer blocks, each of which is composed of a cross-attention (CA) and self-attention (SA) layer.
There are $L=16$ transformer blocks in total.
At each $l$-th transformer layer, a $l$-th CA layer fuses information from the embedding of a conditioning text prompt $\bm{\tau}(y)\in\mathbb{R}^{P\times d_\tau}$ and latent features $\vz^l_t\in\mathbb{R}^{H_l\times W_l\times d_z}$ from a previous layer, aiding in generating images that correspond to a conditioning text prompt.
During this process, a $l$-th CA layer produces a feature map (CA map) $\mathcal{A}_{l,t}^\text{CA}\in\mathbb{R}^{H_l\times W_l \times P}$ at each time step $t$ as follows:
\begin{equation}
\mathcal{A}_{l,t}^\text{CA}=\text{softmax}\left(\frac{Q_{l,t}^{z}\cdot {K_{l}^{\tau}}^\top}{\sqrt{d_l}}\right),
\end{equation}
where $Q_{l,t}^z, K^{\tau}_{l}$ are a query matrix of $\vz^l_t$ and a key matrix of $\bm{\tau}(y)$ obtained via learned linear projections, respectively; and $d_l$ is the dimension of latent features at $l$-th layer.
On the other hand, a $l$-th SA layer is used to learn spatial similarities and correspondences within latent features $\vz^l_t$ from a previous layer.
Similarly, self-attention process produces a SA map $\mathcal{A}_{l,t}^\text{SA}\in\mathbb{R}^{H_l\times W_l \times H_l\times W_l}$ via
\begin{equation}
\mathcal{A}_{l,t}^\text{SA}=\text{softmax}\left(\frac{Q_{l,t}^{z}\cdot {K_{l,t}^{z}}^\top}{\sqrt{d_l}}\right),
\end{equation}
where $Q_{l,t}^z$, $K_{l,t}^{z}$ are query and key matrices of $\vz^l_t$ via learned linear projections.  

Since Stable Diffusion employs U-Net, attention layers in encoders and decoders produce attention maps of different resolutions.
Specifically, attention maps have four different resolutions in total: $(H_l, W_l)\in\{(s_r,  s_r)\}^{3}_{r=0}$, where $s_r\in\{8,16,32,64\}$.
While there are numerous attention maps generated across all diffusion time steps $t\in[1,T]$ and layers $l\in[1,16]$, the fact that there are only four different resolutions suggests that attention maps can be grouped together according to resolution.

\subsection{Aggregation of Attention Maps}
To summarize information from a myriad of generated attention maps, attention maps can be grouped according to resolution and then aggregated via averaging and normalization to have values between 0 and 1~\cite{diffumask,Dataset-Diffusion}:

\begin{equation}
    \bar{\gA}_{s_r}=\frac{1}{\left|\mathbb{L}_{s_r}\right|\cdot T}\sum_{l\in \mathbb{L}_{s_r}}\sum_{1\le t\le T}\frac{\mathcal{A}_{l,t}}{\max(\mathcal{A}_{l,t})},
\end{equation}
where $\mathbb{L}_{s_r}$ is the set of indices of layers that produce attention maps of resolution with scale $s_r$; $\bar{\gA}_{s_r}$ is an aggregated attention map of resolution with scale $s_r$; and $\mathcal{A}_{l,t}$ is either CA map or SA map.
Thus, $\bar{\gA}^\text{SA}_{s_r}$ and $\bar{\gA}^\text{CA}_{s_r}$ represents an aggregated SA map and aggregated CA map of resolution $s_r$, respectively.
Previous works have solely used $\bar{\gA}^\text{CA}_{s_r}$ or its simple multiplication with $\bar{\gA}^\text{SA}_{s_r}$ to find regions corresponding to text in the course of generating a mask.
Hereafter, we will refer to $\bar{\gA}^\text{CA}_{s_k}$ and $\bar{\gA}^\text{SA}_{s_k}$ as CA maps and SA maps, respectively.

\section{Proposed Method}
\begin{figure}[t]
\centering
\includegraphics[width=1\columnwidth]{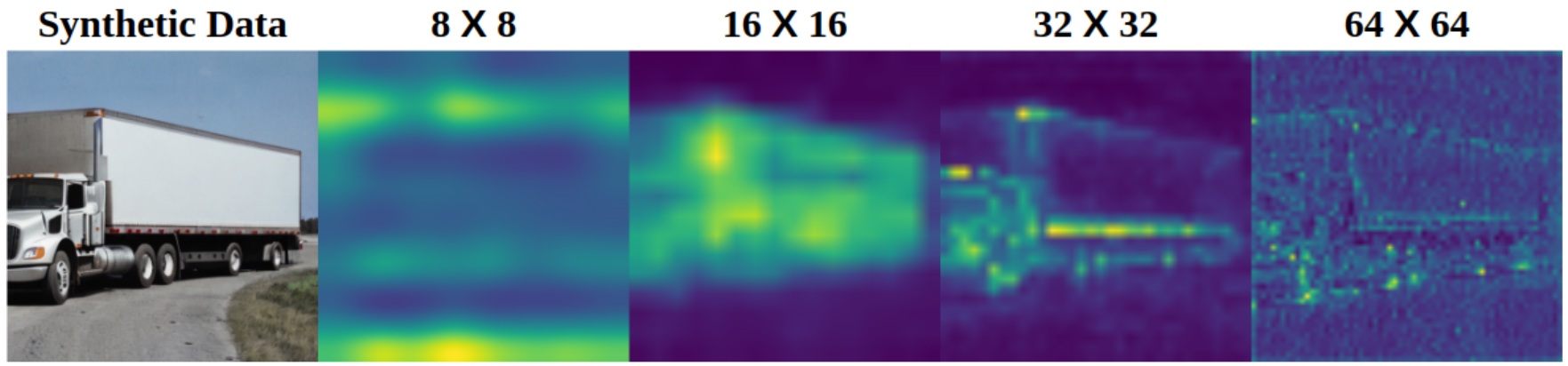} 
\caption{Cross-attention (CA) map at different scales. 
}
\label{fig:CA_map}
\end{figure}
In this work, we aim to extract different yet complementary information out of both cross-attention (CA) and self-attention (SA) maps for the purpose of generating high-quality masks corresponding to generated images and text.
We use CA maps as initial cues (i.e., seeds) to the location of the object at seed intialization.
Seeds, in turn, are then used to expand regions by selecting SA maps that highlight regions similar to the seeds at region expansion process.
To capture fine-grained details, we first attend to whole object and then fine-grained details by iteratively extracting seed from a mask from a previous step (lower resolution) and expanding regions using SA maps of higher resolution.
We further refine a mask by toning down the attention on background and strengthening attention on foreground by using background masks obtained by extracting background seeds expanded to the whole background via SA region expansion, as delineated at mask refinement using background mask process.

\begin{figure}[t]
\centering
\includegraphics[width=1\columnwidth]{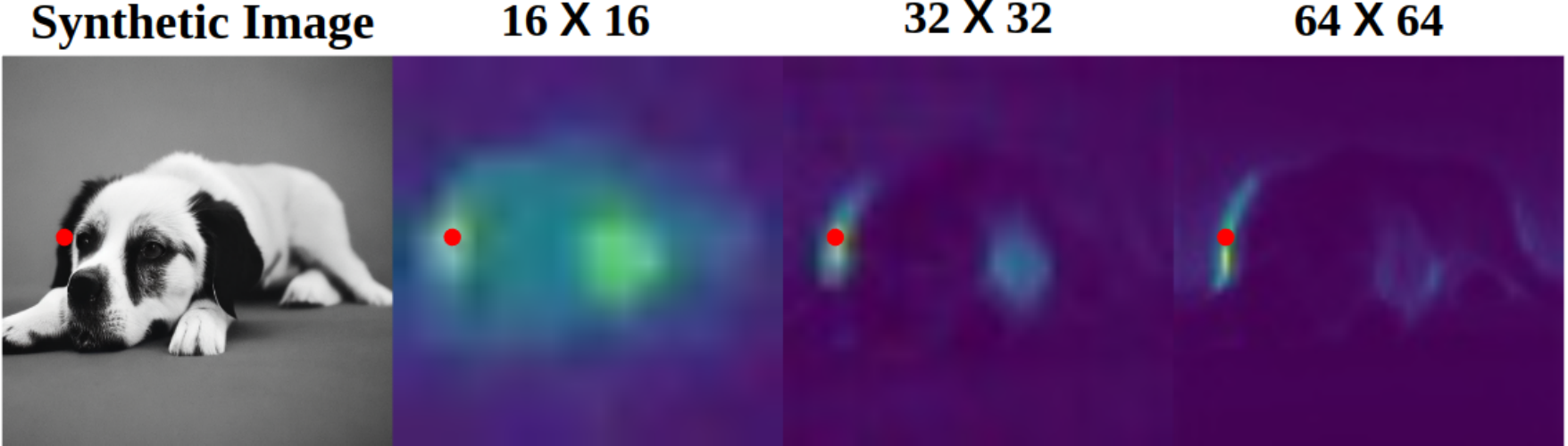} 
\caption{Self-attention (SA) maps at different scales. Red dot indicates the coordinates the visualized multiscale SA maps correspond to.
}
\label{fig:SA_map}
\end{figure}

\subsection{Seed Initialization with Cross-Attention Map}
\label{sec:seed_init}
We start with reiterating that CA maps quantify how similar latent features $\vz$ are to a text embedding $\tau(y)$.
In particular, after simplifying a notation without loss of generality, $\bar{\gA}^\text{CA}[i,j,c]$ quantifies the similarity between a latent vector $\vz$ at $(i,j)$ (i.e., $\vz[i,j]$) and the embedding of $c$-th token in a text prompt $y$ (i.e., $\bm{\tau}(y)[c]$). 
Thus, we can obtain a CA map that highlights regions that correspond to a class of interest via  $\bar{\gA}^\text{CA}[:,:,c_\text{cls}]$ (or $\bar{\gA}^\text{CA}_{c_\text{cls}}$), where $c_\text{cls}$ is the index of a class token in the text prompt.
Indeed, Fig.~\ref{fig:CA_map} shows how a CA map $\bar{\gA}^\text{CA}_{s_r,c_\text{cls}}$ focuses on an object of interest across all scales.
However, the figure exhibits different characteristics across resolution scales.
While a CA map becomes more fine-grained as resolution increases, a CA map tends to attend more uniformly across the whole image and focus less on the object.
On the other hand, a lower-resolution CA map may focus better on the object but provides very coarse semantic correspondence information, losing a lot of details on the structure of the object.
In order to strike a balance between details and precision, we select a $16\times16$ CA map (i.e., $\bar{\gA}^\text{CA}_{16,c_\text{cls}}$) to localize the object of interest.
However, we observe that a $16\times16$ CA map still emphasizes only few specific parts of an object and lacks sharp boundaries, making it inappropriate to be used as final mask.
Upon observation, we propose to obtain initial cues (i.e., seeds) as to the location of an object from a $16\times16$ CA map by collecting spatial coordinates at which a CA value is larger than a threshold $\alpha$ as follows:
\begin{equation}
    \mathbb{S}_1 = \{(i,j) \mid \bar{\gA}^\text{CA}_{16,c_\text{cls}}[i,j]\ge\alpha\}.
\end{equation}

\subsection{Iterative Seed Extraction and Region Expansion with Self-Attention Maps}
\label{sec:region_expand}
While the obtained seeds may locate the object, these seeds attend to only a small portion of it. 
Hinged on the seeds, to expand the attention to the whole object, we turn our attention to the effectiveness of self-attention in learning the similarities between features.
In particular, we note that multiscale SA maps have different levels of attention details at each scale.
Fig.~\ref{fig:SA_map} demonstrates that lower-resolution SA maps attend to broad aspects of the object, while higher-resolution SA maps attend to fine-grained parts of the object.
This observation motivates us to formulate an iterative region expansion, where we first expand to the broad aspects and then fine-grained aspects of the objects.

Formally, with a simplified notation without loss of generality, $\bar{\gA}^\text{SA}[i,j,:,:]$ quantifies the correlation of features $\vz$ between $(i,j)$ and all locations. 
This means that we can attend to other parts of the object by simply indexing $\bar{\gA}^\text{SA}$ with coordinates of seeds:
\begin{equation}
\label{eq:region_expand}
    \gM'_{k} = \frac{1}{\left|\mathbb{S}_{k}\right|}\sum_{(i,j)\in\mathbb{S}_{k}}\bar{\gA}^\text{SA}_{s_k}[i,j,:,:].
\end{equation}
Then, we upsample an obtained mask to a higher resolution $s_{k+1}$, such that we can perform a similar process with SA maps of higher resolution $\bar{\gA}^\text{SA}_{s_{k+1}}$:
\begin{equation}
\label{eq:upsampling}
     \gM_{k+1} = \text{bilinear-upsample}_{s_{k+1}}(\gM'_k),
\end{equation}
where $\text{bilinear-upsample}_{s_{k+1}}$ performs upsampling to $s_{k+1}$ resolution via bilinear-interpolation.

Since we have already located the object, we do not utilize CA maps at higher resolution, which has been shown to give imprecise attention (i.e., attend to background).
Instead, we extract seeds from $\gM_{k+1}$:
\begin{equation}
\label{eq:seed_extraction}
    \mathbb{S}_{k+1} = \{(i,j) \mid {\gM}_{k+1}[i,j]\ge\alpha\}.
\end{equation}
Then, we repeat the process of a mask expansion and refinement using Eq.~\ref{eq:region_expand},~\ref{eq:upsampling},~\ref{eq:seed_extraction} until the last iteration $K$, where the highest resolution of SA maps is reached (i.e., $s_K=64$).
In the last iteration, we only perform region expansion and do not perform upsampling and seed extraction.

\begin{figure}[t]
\centering
\includegraphics[width=1\columnwidth]{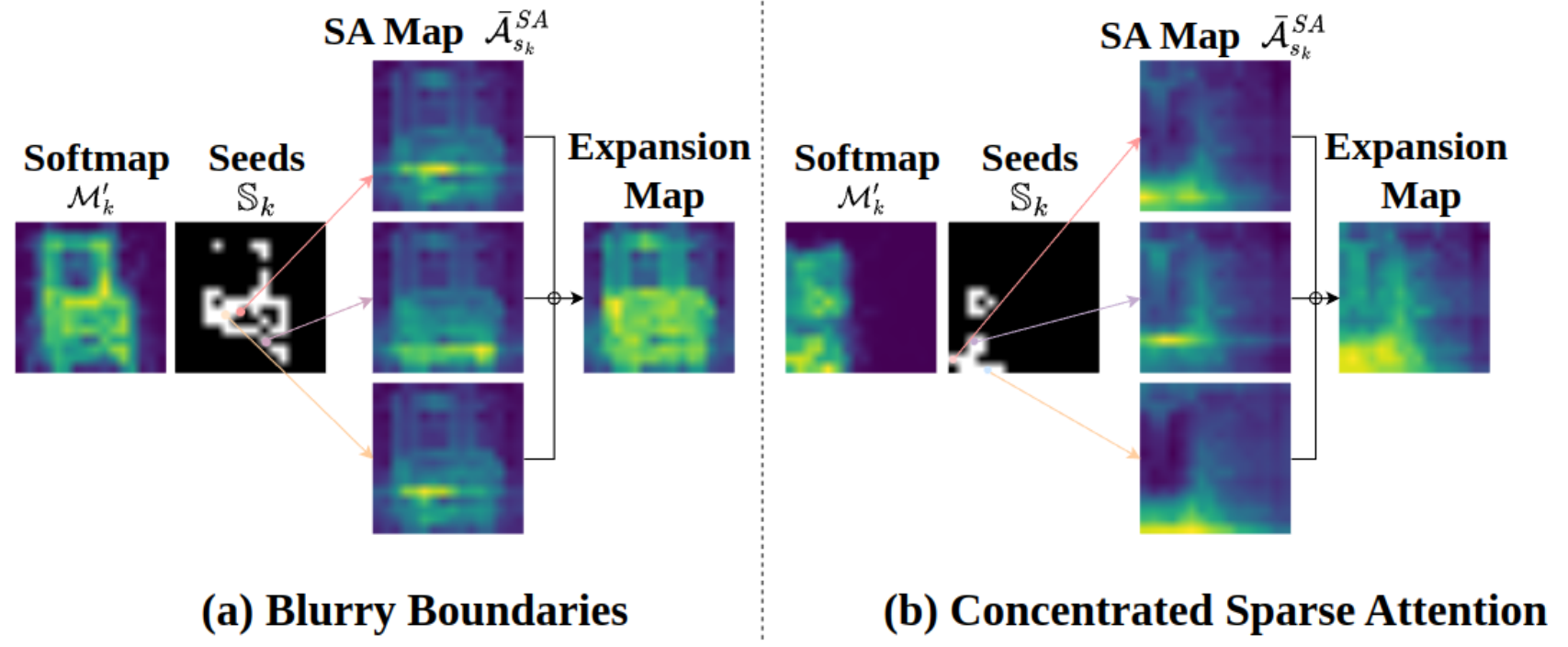} 
\caption{Persistent errors from CA map seeds. (a) CA map seeds can point to background region. (b) CA map seeds can be sparse and concentrated.}
\label{fig:error_prop}
\end{figure}

\subsection{Mask Refinement with Background Mask}
\label{sec:bg_att}
While the iterative region expansion process with SA Maps has substantially enhanced the quality of a mask, we observe that initial seeds from CA maps introduce two persistent errors that propagate throughout the process, as illustrated in Fig.~\ref{fig:error_prop}.
The first error is that the initial seeds sometimes point at background region near the object boundary, especially when objects have detailed and complex structures.
These inaccurate seeds result in region expansion to nearby background regions, thereby creating blurry boundary after region expansion with SA, as depicted in Fig.~\ref{fig:error_prop}(a).
The second error is that seeds are often sparse and concentrated at the certain parts of an object with non-homogeneous appearances, as CA finds strong correlation between certain parts of an object and a text.
When seeds are sparse and concentrated and features of objects have large variances, region expansion with SA maps fails to resolve the issue, as shown in Fig.~\ref{fig:error_prop}(b).

\begin{figure}[t]
\centering
\includegraphics[width=1\columnwidth]{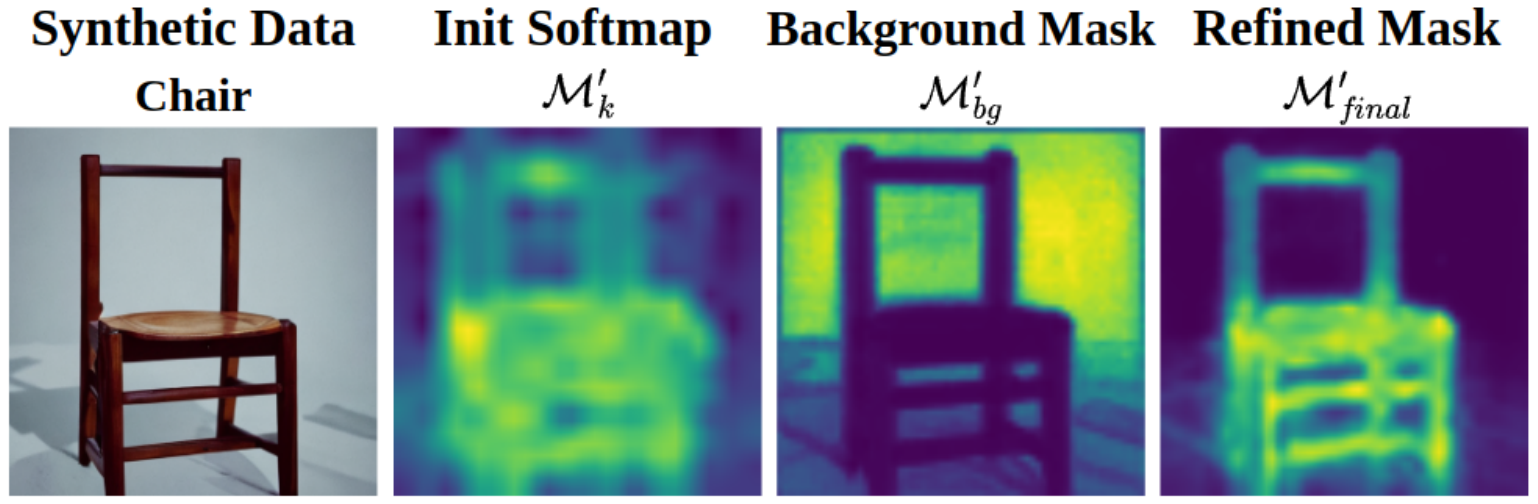} 
\caption{Mask refinement with a background mask.}
\label{fig:be_effect}
\end{figure}

\begin{table*}[]
\centering
\resizebox{\textwidth}{!}{%
\begin{tabular}{@{}lccccccccccccccc@{}}
\toprule
\multicolumn{1}{l|}{} & \multicolumn{1}{c|}{} & \multicolumn{1}{c|}{} & \multicolumn{2}{c|}{Prompt Tuning} & \multicolumn{10}{c|}{Semantic Segmentation (IoU) for Selected Classes (\%)} & \multicolumn{1}{c}{mIoU (\%)} \\
\multicolumn{1}{l|}{Train Set} & \multicolumn{1}{c|}{Number} & \multicolumn{1}{c|}{Backbone} & {LLM} & \multicolumn{1}{c|}{Retrieval} &  aeroplane & bus & cat & chair & cow & dog & horse & person & sheep & \multicolumn{1}{c|}{sofa} & val\\ \midrule
\midrule
\multicolumn{16}{l}{\textit{Training with Pure Real Data}} \\
\multicolumn{1}{l|}{\multirow{3}{*}{VOC}} & \multicolumn{1}{c|}{R: 10.6k (all)} & \multicolumn{1}{c|}{ResNet50} & \xm &  \multicolumn{1}{c|}{\xm} & 87.5 & 95.5 & 92.2 & 44.0 & 85.4 & 89.1 & 82.1 & 89.2 & 80.6 & \multicolumn{1}{c|}{53.6} & 77.3 \\
\multicolumn{1}{l|}{} & \multicolumn{1}{c|}{R: 10.6k (all)} & \multicolumn{1}{c|}{Swin-B} & \xm & \multicolumn{1}{c|}{\xm} & 97.0 & 91.7 & 96.5 & 57.5 & 95.9 & 96.8 & 94.4 & 92.5 & 95.1 & \multicolumn{1}{c|}{65.6} & 84.3 \\
\multicolumn{1}{l|}{} & \multicolumn{1}{c|}{R: 5.0k} & \multicolumn{1}{c|}{Swin-B} & \xm & \multicolumn{1}{c|}{\xm} & 95.5 & 96.1 & 95.2 & 47.3 & 90.3 & 92.8 & 94.6 & 90.9 & 93.7 & \multicolumn{1}{c|}{61.4} & 83.4 \\ \midrule
\multicolumn{16}{l}{\textit{Training with \textbf{Pure Synthetic Data}}} \\
\multicolumn{1}{l|}{Dataset Diffusion}  & \multicolumn{1}{c|}{S: 40.0k} & \multicolumn{1}{c|}{ResNet50} & \cm & \multicolumn{1}{c|}{\cm} & - & - & - & - & - & - & - & - & - & \multicolumn{1}{c|}{-} & 60.4 \\
\midrule
\multicolumn{1}{l|}{\multirow{2}{*}{DiffuMask}}  & \multicolumn{1}{c|}{S: 60.0k} & \multicolumn{1}{c|}{ResNet50} & \xm & \multicolumn{1}{c|}{\cm} & 80.7 & 81.2 & 79.3 & 14.7 & 63.4 & 65.1 & 64.6 & 71.0 & 64.7 & \multicolumn{1}{c|}{27.8} & 57.4 \\
\multicolumn{1}{l|}{} & \multicolumn{1}{c|}{S: 60.0k} & \multicolumn{1}{c|}{Swin-B} & \xm & \multicolumn{1}{c|}{\cm} & 90.8 & 88.3 & 92.5 & 27.2 & 92.2 & 86 & 89 & 76.5 & 92.2 & \multicolumn{1}{c|}{49.8} & 70.6 \\ \midrule

\multicolumn{1}{l|}{\multirow{4}{*}{\textbf{SeeDiff (Ours)} }} & \multicolumn{1}{c|}{S: 40.0k} & \multicolumn{1}{c|}{ResNet50} & \xm  & \multicolumn{1}{c|}{\xm} & 81.6 & 75.8 & 81.8 & 13.2 & 59.0 & 62.5 & 59.3 & 61.6 & 70.2 & \multicolumn{1}{c|}{48.5} & 61.2 \\
\multicolumn{1}{l|}{} & \multicolumn{1}{c|}{S: 40.0k} & \multicolumn{1}{c|}{Swin-B} & \xm & \multicolumn{1}{c|}{\xm} & 94.8 & 93.1 & 88.0 & 22.9 & 90.9 & 87.3 & 71.9 & 73.0 & 94.7 & \multicolumn{1}{c|}{63.1} & 78.6 \\
\multicolumn{1}{l|}{} & \multicolumn{1}{c|}{S: 60.0k} & \multicolumn{1}{c|}{ResNet50} & \xm &\multicolumn{1}{c|}{\xm} & 82.3 & 75.6 & 82.7 & 15.6 & 62.2 & 61.7 & 64.1 & 67.5 & 73.0 & \multicolumn{1}{c|}{51.2} & 62.6 \\
\multicolumn{1}{l|}{} & \multicolumn{1}{c|}{S: 60.0k} & \multicolumn{1}{c|}{Swin-B} & \xm & \multicolumn{1}{c|}{\xm} & 95.5 & 92.0 & 91.5 & 22.4 & 91.5 & 85.1 & 78.3 & 74.9 & 94.2 & \multicolumn{1}{c|}{65.3} & \textbf{79.6} \\
\midrule
\multicolumn{16}{l}{\textit{Finetune on \textbf{Real Data}}} \\
\multicolumn{1}{l|}{\multirow{2}{*}{DiffuMask}}  & \multicolumn{1}{c|}{S: 60.0k+R: 5.0k} & \multicolumn{1}{c|}{ResNet50} & \xm & \multicolumn{1}{c|}{\cm} & 85.4 & 92.9 & 91.7 & 38.4 & 86.5 & 86.2 & 82.5 & 87.5 & 81.2 & \multicolumn{1}{c|}{39.8} & 77.6 \\
\multicolumn{1}{l|}{} & \multicolumn{1}{c|}{S: 60.0k+R: 5.0k} & \multicolumn{1}{c|}{Swin-B} & \xm & \multicolumn{1}{c|}{\cm} & 95.6 & 96.9 & 96.6 & 51.5 & 96.7 & 95.5 & 96.1 & 91.5 & 96.4 & \multicolumn{1}{c|}{70.2} & 84.9 \\
\midrule
\multicolumn{1}{l|}{\multirow{2}{*}{\textbf{SeeDiff (Ours)} }} & \multicolumn{1}{c|}{S: 60.0k+R: 5.0k} & \multicolumn{1}{c|}{ResNet50} & \xm & \multicolumn{1}{c|}{\xm} & 91.9 & 82.0 & 93.7 & 25.7 & 67.9 & 77.0 & 74.6 & 76.6 & 73.8 & \multicolumn{1}{c|}{56.0} & 71.4 \\
\multicolumn{1}{l|}{} & \multicolumn{1}{c|}{S: 60.0k+R: 5.0k} & \multicolumn{1}{c|}{Swin-B} & \xm &\multicolumn{1}{c|}{\xm} & 98.5  & 98.6 & 96.7 & 49.9 & 89.5 & 91.8 & 98.4 & 92.0 & 97.4 &  \multicolumn{1}{c|}{67.8} & \textbf{88.8} \\
\bottomrule
\end{tabular}%
}
\caption{Semantic segmentation results on VOC 2012 \texttt{val}. 
 `S' and `R' denote synthetic images and real images, respectively.
}
\label{table_1}
\end{table*}

To resolve these two issues, we start with an observation that a synthetic image generated by Stable Diffusion with a simple text prompt generates a relatively uniform background.
Upon our observation, we claim that it is much easier for SA maps to give uniformly distributed attention to background.
This means we can refine a mask of class by down-weighting background parts and up-weighting class regions, via subtraction of an accurate background mask from a mask of class.
To obtain a background mask, we simply inverse a mask obtained from the last iteration of the region expansion process in region expansion:
\begin{equation}
     \gM'_\text{inv} = (1-\gM'_K),
\end{equation}
from which background seeds $\mathbb{S}_\text{bg}$ are obtained, similar to Eq.~\ref{eq:seed_extraction}.
Then, a background mask $\gM'_\text{bg}$ is finally obtained from $\mathbb{S}_\text{bg}$ via region expansion with SA maps (Eq.~\ref{eq:region_expand}).
Then, we obtain a more refined mask by multiplying a class-region mask $\gM'_K$ with the inverse of a background mask $\gM'_\text{bg}$, such that background regions are diminished while class regions are amplified:
\begin{equation}
     \gM'_\text{final} = (1-\gM'_\text{bg})\odot\gM'_K,
\end{equation}
which is then upsampled to full resolution $512\times512$ to produce a higher resolution mask $\gM_\text{final}$, followed by binarization with a threshold $\beta$ to obtain a final annotation mask $\vm$ for a synthetic image $\vx$:
\begin{equation}
     \vm[i,j] = \gM_\text{final}[i,j]\odot\mathds{1}_{\{\gM_\text{final}[i,j]\ge\beta\}},
\end{equation}
where $\mathds{1}_{\{\cdot\}}$ is an indicator function which gives $1$ when the condition in subscript is met and $0$ otherwise.

\section{Experiments}
\label{sec:experiments}

\subsection{Experimental Settings}
\label{sec:experiments setup} 
\subsubsection{Datasets.} Following the settings of the previous work DiffuMask~\cite{diffumask}, we evaluated our model on the following two datasets Pascal-VOC2012~\cite{pascal} and Cityscapes~\cite{Cityscapes}.
In particular, we also conducted an evaluation on the Pascal-VOC2012 dataset in the open-vocabulary semantic segmentation setting, where the evaluation is conducted by dividing into 15 seen classes and 5 unseen classes, following the setup of the previous model~\cite{Zegformer,diffumask}.

\subsubsection{Evaluation Metric.} We use mIoU score as a metric to evaluate the segmentation task, which calculates the intersection between the predicted regions and the actual ground truth masks, and then takes the average value.

\subsubsection{Implementation Details.} We construct our components using a pre-trained Stable Diffusion~\cite{LDM} model. 
We utilize the Stable Diffusion 2-base version to generate images with $T=50$ timesteps as denoising step. 
We use only the pre-trained internal features without any fine-tuning of the Stable Diffusion model.
As for text prompts, we generate the dataset using the widely used text prompt in the form of `a photo of a $<\texttt{CLASS}>$'.
We utilize $\alpha=0.5$ as a threshold parameter to extract the seeds and $\beta=0.3$ as a threshold parameter to discretize a soft mask to a final mask.
Following previous works~\cite{diffumask,Dataset-Diffusion}, to evaluate the generated images and masks, we train and evaluate the segmentation model Mask2Former~\cite{Mask2Former}. 
The settings required for training and evaluating Mask2Former, including initialization, data augmentation, batch size, weight decay, and learning rate, are configured according to the original paper.
In the PascalVOC-2012~\cite{pascal} setting, we generate 2$k$ and 3$k$ images per class, using a total number of images (40.0$k$ and 60.0$k$) identical to those used in previous studies such as Diffusion Dataset~\cite{Dataset-Diffusion} and DiffuMask~\cite{diffumask}. 
For data augmentation, we apply the same techniques used in DiffuMask, including splicing, gaussian noise, perspective transformation, and occlusion. 
In the Cityscapes dataset, we conduct evaluations only on the `Human' and `Vehicle' categories, following the same evaluation setting as same as our baseline DiffuMask~\cite{diffumask}. 
The `Human' and `Vehicle' categories consist of six sub-classes: person, rider, car, bus, truck, and train.
All experiments, including image generation and evaluation, have been conducted on NVIDIA 4090 RTX GPU.
Since we utilize only the internal parameters of Stable Diffusion, there is no significant increase in computational cost.

\subsection{Experimental Results}
\label{sec:experiments setup} 

\subsubsection{Semantic Segmentation on VOC 2012.} Table~\ref{table_1} shows the results of semantic segmentation on the VOC 2012 dataset. 
We generate 40k and 60k images to match the number of generated images for a fair comparison with previous work.~\cite{diffumask,Dataset-Diffusion}
Training Mask2Former~\cite{Mask2Former} with our generated data results in significant performance improvement.
Additional fine-tuning with real data brings further improvement, surpassing the performance of a model (Swin-B backbone) trained solely with the whole real data by a large margin ($4\sim5\%$).
We also note that our generated images bring substantial performance improvements, compared to baselines, without any prompt tuning.
Also, even when using less number of images, SeeDiff outperforms DiffuMask.
The results underline the effectiveness of our method SeeDiff in generating high-quality masks.
\subsubsection{Cityscapes.}
Table~\ref{table_2} displays the results of semantic segmentation on Cityscapes, where similar tendencies are observed.
Again for Cityscapes dataset, we use input text with fixed templates.
Considering the domain of Cityscapes, which consists of urban street scenes, we use the fixed prompt `a photo of a $<\texttt{CLASS}>$ in the city.'

\begin{table}[t]
\centering
\resizebox{.4\textwidth}{!}{ %
\small
\begin{tabular}{@{}lcccc@{}}
\toprule
\multicolumn{4}{c}{Module} & \multicolumn{1}{|c}{mIoU (\%) } 
\\
\multicolumn{1}{l|}{\textbf{CAA}} & \multicolumn{1}{c|}{\textbf{CRE}} & \multicolumn{1}{c|}{\textbf{IRE}} & \multicolumn{1}{c|}{\textbf{BE}} & \texttt{VOC\textsubscript{val}}
\\
\midrule \midrule
\multicolumn{1}{l|}{\ding{51}}  & \multicolumn{1}{c|}{\ding{55}} & \multicolumn{1}{c|}{\ding{55}} & \multicolumn{1}{c|}{\ding{55}} & 68.8\\
\multicolumn{1}{l|}{\ding{51}}  & \multicolumn{1}{c|}{\ding{51}} & \multicolumn{1}{c|}{\ding{55}} & \multicolumn{1}{c|}{\ding{55}}& 71.0\\
\multicolumn{1}{l|}{\ding{51}}  & \multicolumn{1}{c|}{\ding{51}} & \multicolumn{1}{c|}{\ding{51}} & \multicolumn{1}{c|}{\ding{55}} & 73.2\\
\multicolumn{1}{l|}{\ding{51}}  & \multicolumn{1}{c|}{\ding{51}} & \multicolumn{1}{c|}{\ding{51}} & \multicolumn{1}{c|}{\ding{51}} & \textbf{79.6}\\
\midrule
\end{tabular} %
}
\caption{Module Ablations. We perform ablations of our module with \mbox{VOC 2012 val}, using Mask2Former with Swin-B. CAA denotes cross-attention aggregation, representing our baseline.
CRE denotes cross-attention guided region expansion. 
IRE denotes iterative region expansion.
BE denotes background expansion inversion. }
\label{module_table}
\end{table}

\begin{table}[t]
\centering
\resizebox{.4\textwidth}{!}{ %
\scriptsize
\Large
\begin{tabular}{@{}lcccc@{}}
\toprule
\multicolumn{1}{c|}{} & \multicolumn{1}{c|}{} & \multicolumn{1}{c|}{} & \multicolumn{2}{c}{IoU\% } 

\\
\multicolumn{1}{l|}{Train Set} & \multicolumn{1}{c|}{Number} & \multicolumn{1}{c|}{Backbone} & Human & \multicolumn{1}{c}{Vehicle} \\ 
\midrule \midrule
\multicolumn{5}{l}{\textit{Train with Pure Real Data}} \\
\multicolumn{1}{l|}{\multirow{2}{*}{Cityscapes}} & \multicolumn{1}{c|}{3.0k (all)} & \multicolumn{1}{c|}{Swin-B} & 85.5 & 96.0   \\
\multicolumn{1}{l|}{} & \multicolumn{1}{c|}{1.5k} & \multicolumn{1}{c|}{Swin-B} & 84.6 & 95.3 \\
\midrule
\multicolumn{5}{l}{\textit{Train with Pure Synthetic Data}} \\
\multicolumn{1}{l|}{\multirow{1}{*}{DiffuMask}}  & \multicolumn{1}{c|}{100.0k} & \multicolumn{1}{c|}{Swin-B} & 72.1 & 87.0 \\
\midrule
\multicolumn{1}{l|}{\multirow{1}{*}{\textbf{SeeDiff (ours)}}}  & 
\multicolumn{1}{c|}{100.0k} & \multicolumn{1}{c|}{Swin-B} & 74.3 & 89.8   \\
\bottomrule
\end{tabular} %
}
\caption{{Result of Semantic Segmentation on Cityscapes \mbox{val.}} }
\label{table_2}
\end{table}
\subsubsection{Zero-Shot Semantic Segmentation.} 
In the zero-shot setting of VOC 2012, our model outperform DiffuMask and several models trained with real data.
Notably, we achieve similar performance on `seen' and `unseen' categories, while DiffuMask has a drop in performance on `unseen categories.
We believe this is because DiffuMask uses a pre-trained segmentation network, which limits the performance on classes unseen during the pre-training of a segmentation network.

\subsection{Ablation Study}
In this section, extensive ablation studies are conducted to assess the effectiveness of each proposed module.
We perform ablation studies on Pascal-VOC2012, where 60k images are generated to train Mask2Former (Swin-B).

\subsubsection{Module Ablations.} 
Table~\ref{module_table} assesses the effectiveness of each component of our framework SeeDiff.
In the case of Cross-Attention Aggregation (CAA), the final mask is generated solely from the aggregated cross-attention maps during the denoising step, similar to the mask generation method used by a baseline, DiffuMask~\cite{diffumask}.
Cross-attention guided region expansion (CRE) generates the final mask by aggregating the self-attention maps corresponding to the seeds extracted from the CA map at each resolution of the U-Net.
Iterative region expansion (IRE) denotes a variant of our method, which uses a soft mask $\mathcal{M}_k'$ from low resolution as an initial mask for the next resolution. 
Finally, IRE with mask refinement with background expansion (BE) represents our final model.

\subsubsection{Study on Different Feature Scales.} 
Table~\ref{table_5} evaluates the influence of CA maps of different scales.
Through this experiment, we observe that higher resolutions tend to lose objectness, as also exhibited in Figure~\ref{fig:CA_map}. 
This validates the our design choice of using CA map of $16\times16$ resolution.

\subsubsection{Study on Mask Generation Strategy.} Table~\ref{mask_strategy} compares different mask generation strategies: CA (Cross-Attention-based mask generation; DiffuMask~\cite{diffumask}), CA$\cdot$SA (Cross-Attention-Self-Attention-Multiplication-based mask generation; Dataset Diffusion~\cite{Dataset-Diffusion}), and our SeeDiff.
SeeDiff demonstrates to be most effective.
The results validate our motivation of drawing connections with classical seeded segmentation algorithms.

\begin{table}[t]
    \begin{minipage}{.49\linewidth}
        \centering

\begin{tabular}{@{}lccc@{}}
        \toprule
        \multicolumn{1}{c}{} & \multicolumn{1}{c}{} & \multicolumn{1}{c}{} & \multicolumn{1}{|c}{mIoU (\%) } \\
        \multicolumn{3}{l|}{Resolution} & \texttt{VOC\textsubscript{val}} \\
\midrule \midrule
        \multicolumn{1}{l}{16}  & \multicolumn{1}{c}{} & \multicolumn{1}{c|}{} & \textbf{65.9}\\
        \multicolumn{1}{l}{32}  & \multicolumn{1}{c}{} & \multicolumn{1}{c|}{} & 64.3\\
        \multicolumn{1}{l}{64}  & \multicolumn{1}{c}{} & \multicolumn{1}{c|}{} & 58.2\\
\bottomrule
\end{tabular} %

\caption{Study on the influence of resolution of CA map. We evaluate the influence of resolution of CA map on the final performance.}
    
\label{table_5}

\end{minipage} 
    \hfill
    \begin{minipage}{.49\linewidth}
        \centering
\begin{tabular}{@{}lcc@{}}
\toprule
\multicolumn{1}{c}{} & \multicolumn{1}{c|}{} &  \multicolumn{1}{c}{mIoU (\%) } 
\\
\multicolumn{1}{l}{Method} & \multicolumn{1}{c|}{} &  \multicolumn{1}{c}{\texttt{VOC\textsubscript{val}}} \\
\midrule \midrule
\multicolumn{1}{l}{\multirow{1}{*}{CA}}{}  & \multicolumn{1}{c|}{}  & 68.8\\
\multicolumn{1}{l}{\multirow{1}{*}{CA$\cdot$SA}}  & \multicolumn{1}{c|}{} & 72.0 \\
\multicolumn{1}{l}{\multirow{1}{*}{SeeDiff}}  & \multicolumn{1}{c|}{} &  \textbf{79.6} \\
\bottomrule
\end{tabular} %

\caption{Comparisons of Attention-based Mask Generation Strategy. CA denotes cross-attention aggregation (DiffuMask). CA$\cdot$SA denotes simple multiplication of CA and SA (Dataset Diffusion). }
\label{mask_strategy}
    \end{minipage} 
\end{table}

\section{Conclusion}
In this study, we introduce a novel off-the-shelf framework, named \textbf{SeeDiff}, that generates high-quality image-mask pairs from Stable Diffusion, without any training, prompt tuning, or using pre-trained segmentation networks.
The design of a framework is inspired by a classical seeded segmentation algorithm, which expands initial point cues (i.e., seeds).
We observe that CA maps provide coarse localization of objects, serving as seeds.
In the mean time, we utilize SA map to perform mask region expansion to cover the whole object.
Consequently, SeeDiff generates high-quality image-mask pairs that can be used to train a segmentation network with outstanding performance.

\section{Acknowledgments}
This work was supported by Institute of Information \& communications Technology Planning \& Evaluation (IITP) grant funded by the Korea government(MSIT) (No.RS-2020-II201373, Artificial Intelligence Graduate School Program(Hanyang University)

\bibliography{aaai25,supple}

\newpage
\newpage
\clearpage

\input{supple}

\end{document}

%% file: supple.tex
\setcounter{section}{0}
\setcounter{figure}{0}
\setcounter{table}{0}
\setcounter{equation}{0}

\renewcommand{\thetable}{S\arabic{table}}
\renewcommand{\thesection}{S\arabic{section}}
\renewcommand{\thefigure}{S\arabic{figure}}
\renewcommand{\theequation}{S\arabic{equation}}

\lstset{%
	basicstyle={\footnotesize\ttfamily},
	numbers=left,numberstyle=\footnotesize,xleftmargin=2em,
	aboveskip=0pt,belowskip=0pt,%
	showstringspaces=false,tabsize=2,breaklines=true}
\floatstyle{ruled}
\newfloat{listing}{tb}{lst}{}
\floatname{listing}{Listing}

\pdfinfo{
/TemplateVersion (2025.1)
}
\definecolor{brickred}{rgb}{0.8, 0.25, 0.33}
\definecolor{brickgreen}{rgb}{0.25, 0.8, 0.33}
\setcounter{secnumdepth}{2}

\clearpage
\begin{center}
{\bf {\LARGE Supplementary Material}}
\end{center}

\begin{center}{\bf {\Large SeeDiff: Off-the-Shelf Seeded Mask Generation from Diffusion Models}}
\end{center}

\maketitle

In this supplementary material, we provide additional context and elaboration on the field related to our research in Sec.~\ref{sec:related_works}. 
Then, in Sec.~\ref{sec:additional ablation studies}, we conduct additional experiments for our ablation study. Following the experimental setup of our baseline model, DiffuMask, we performed these additional experiments on various datasets using only basic prompts.

\section{Additional Related Works}
\label{sec:related_works}
\subsection{Seeded Segmentation}
Semantic segmentation aims to distinguish the semantic area of the input image by performing pixel-level classification. 
However, obtaining pixel-level annotations necessary for ensuring model performance is a costly endeavor. 
In contrast, image-level annotations offer a more economical and efficient alternative, though they necessitate the inference of pixel-level object locations due to the absence of explicit object location information.
The spatial continuity of semantic labels within the same object implies that pixels in proximity to a discriminative area typically belong to the same object. 
This characteristic renders seed-based segmentation an effective approach for region delineation through pixel expansion.
Classical methods~\cite{seeded,Seeded-region-growing,Seed-manually2} require manual initialization of the seed, followed by segmentation based on pixel similarity, employing graph theory~\cite{graph,RandomWalk} and energy optimization techniques.
However, these approaches are limited by their dependence on user-defined seed placement and similarity criteria based on low-level features (e.g. color, intensity, or texture), which can lead to over-segmentation and inaccuracies.
To address these limitations, deep learning-based seeded segmentation methods~\cite{weakly-super,seed_2} have been developed. 
By combining the seeded segmentation method and deep learning, the seed cues generated by the classification network is used as the initial seed to prevent incorrect seed placement, and the improved level of image features is used to evaluate the similarity between pixels and enable more sophisticated expressions. 
This allows more accurate segmentation.

\subsection{Semantic Segmentation Using Diffusion Models} 
Diffusion-based models~\cite{LDM} have garnered significant attention in text-to-image generation and have recently been effectively utilized in discriminative computer vision studies.
Diffusion models trained on large-scale datasets like LAION-5B~\cite{LAION-5B} learn detailed semantic information through the U-Net~\cite{Unet} architecture during the image synthesis process and understand the correspondence between visual pixels and language
These models are applied to various segmentation tasks by leveraging their ability to model complex data distributions and utilize internal representations that are associated with high-level semantic concepts during the generation process.

Diffusion models extend the denoising process to generate segmentation mask distributions~\cite{DDp,Semantic-Diffusion,SegDiff} based on image conditions through training. 
Specifically, these models can predict masks for images using a multi-step denoising process~\cite{DiffusionInst,Segmentation-Ensembles}, and Maskdiff~\cite{MaskDiff} models the distribution of masks and reconstructs the mask from a noise-to-filter perspective.
Additionally, the activation at certain diffusion steps effectively captures semantic information~\cite{label-efficient}, enabling strong performance in few-shot semantic segmentation.
Many studies also focus on extracting internal representations from pretrained diffusion models, as these representations are well-distinguished and interrelated with high/mid-level semantic concepts that can be described in language.
Combining the representational power of diffusion models with the classification capabilities of discriminative models~\cite{panoptic-segmentation} or analyzing the influence of visual and linguistic information on pixel-space relationships~\cite{DAAM}.
By leveraging the features of diffusion models, you can apply them to semantic segmentation in various ways.
However, these models still require additional training and Ground Truth data.
In contrast, our approach utilizes a pretrained Stable Diffusion model to perform semantic segmentation using the internal information of the diffusion model, without the need for additional training or ground truth data.

\section{Additional Ablation Studies}
\label{sec:additional ablation studies}

\subsection{Zero-Shot Semantic Segmentation}
\begin{table}[t]
\centering
\resizebox{.49\textwidth}{!}{ %
\scriptsize
\Large
\begin{tabular}{@{}lccccc@{}}
\toprule
\multicolumn{1}{c|}{} & \multicolumn{2}{c|}{Train Set} & \multicolumn{3}{c}{mIoU (\%)} \\
\multicolumn{1}{l|}{Methods} & Type & \multicolumn{1}{c|}{Categories} & Seen & Unseen & Harmonic \\ \midrule
\multicolumn{6}{l}{Manual Mask Supervision} \\
\multicolumn{1}{l|}{ZS3} & real & \multicolumn{1}{c|}{15} & 78.0 & 21.2 & 33.3\\
\multicolumn{1}{l|}{CaGNet} & real & \multicolumn{1}{c|}{15} & 78.6 & 30.3 & 43.7
\\
\multicolumn{1}{l|}{Joint} & real & \multicolumn{1}{c|}{15} &77.7 &32.5 &45.9  \\
\multicolumn{1}{l|}{STRICT} & real & \multicolumn{1}{c|}{15} &82.7 &35.6 &49.8 \\
\multicolumn{1}{l|}{SIGN} & real & \multicolumn{1}{c|}{15} &83.5 &41.3 &55.3  \\
\multicolumn{1}{l|}{ZegFormer } & real & \multicolumn{1}{c|}{15} &86.4 &63.6 &73.3  \\ 
\midrule
\multicolumn{6}{l}{\textit{Image and Caption only, pre-trained on CC-12M, CC-3M}} \\
\multicolumn{1}{l|}{TCL } & real & \multicolumn{1}{c|}{15+5} &- &- &83.3  \\ 
\midrule

\multicolumn{6}{l}{\textit{Pseudo Mask Supervision from Model pre-trained on COCO}} \\
\multicolumn{1}{l|}{Li et al. {[}36{]} (ResNet101)} & synthetic & \multicolumn{1}{c|}{15+5} &62.8 &50.0 &55.7  \\ \midrule

\multicolumn{6}{l}{\textit{Text(Prompt) Supervision}} \\
\multicolumn{1}{l|}{DiffuMask (R50)} & synthetic & \multicolumn{1}{c|}{15+5} &60.8 &50.4 &55.1  \\
\multicolumn{1}{l|}{DiffuMask (Swin-B)} & synthetic & \multicolumn{1}{c|}{15+5} &71.4 &65.0 &68.1 \\ \midrule
\multicolumn{1}{l|}{\textbf{SeeDiff (Ours, R50)}} & synthetic & \multicolumn{1}{c|}{15+5} & 63.4 & 61.5 & 62.4 \\
\multicolumn{1}{l|}{\textbf{SeeDiff (Ours, Swin-B)}} & synthetic & \multicolumn{1}{c|}{15+5} & 79.9 & \textbf{78.6} & 79.3\\

\midrule

\end{tabular} %
}
\caption{\textbf{Performance for Zero-Shot Semantic Segmentation Task on PASCAL VOC}}
\label{table:zero-shot-voc}
\end{table}

We provide the experimental results on Table~\ref{table:zero-shot-voc} for the VOC dataset, where the training and evaluation were conducted in a zero-shot setting with 15 seen classes and 5 unseen classes. 
The results include the seen mIoU, unseen mIoU, and harmonic mIoU.

\subsection{Additional Evaluation} 
We evaluate the performance on other dataset ADE-20$k$, using only the basic prompt in the form of `a photo of a $<\texttt{CLASS}>$', without any prompt tuning or utilizing domain knowledge of the datasets. 
Our goal is to demonstrate that synthetic images and masks can be effectively used under these conditions.

\begin{table}[t]
\centering
\resizebox{.49\textwidth}{!}{ %
\begin{tabular}{@{}lccccc@{}}
\toprule
\multicolumn{1}{l}{Train Set} & \multicolumn{1}{|c}{Test set} & \multicolumn{1}{|c}{Car} & \multicolumn{1}{c}{Person} & \multicolumn{1}{c}{Motorbike} & \multicolumn{1}{|c}{mIoU ($\%$)}
\\
\midrule
\multicolumn{1}{l|}{DiffuMask} & \multicolumn{1}{c|}{VOC 2012 \texttt{val}} &   \multicolumn{1}{c}{74.2} &  \multicolumn{1}{c}{71.0} &  \multicolumn{1}{c|}{63.2} &  \multicolumn{1}{c}{69.5}\\
\multicolumn{1}{l|}{\textbf{SeeDiff (ours)}} & \multicolumn{1}{c|}{VOC 2012 \texttt{val}} &  \multicolumn{1}{|c}{75.9} & \multicolumn{1}{c}{67.5} & \multicolumn{1}{c|}{66.3} & \multicolumn{1}{c}{69.9}\\
\midrule %
\multicolumn{1}{l|}{DiffuMask} & \multicolumn{1}{c|}{Cityscapes \texttt{val}} &   \multicolumn{1}{c}{84.0} & \multicolumn{1}{c}{70.7} & \multicolumn{1}{c|}{23.6} & \multicolumn{1}{c}{59.4}\\
\multicolumn{1}{l|}{\textbf{SeeDiff (ours)}} & \multicolumn{1}{c|}{Cityscapes \texttt{val}} & \multicolumn{1}{c}{86.4} & \multicolumn{1}{c}{67.1} &  \multicolumn{1}{c|}{28.2} & \multicolumn{1}{c}{60.5}\\
\midrule
\end{tabular}
}
\caption{\textbf{Performance for domain generalization between different datasets.} We use the basic prompt `a photo of a $<\texttt{CLASS}>$' to generate images and pixel-level annotations. 
We performed evaluations on different datasets using the same generated dataset. }
\label{table:domain_generalization}
\end{table}

\begin{table}[t]
\centering
\resizebox{.49\textwidth}{!}{ %
\small
\begin{tabular}{@{}lcccccc@{}}
\toprule
\multicolumn{1}{l}{Train Set} & \multicolumn{1}{|c}{Number} & \multicolumn{1}{|c}{Backbone} & \multicolumn{1}{|c}{bus} & \multicolumn{1}{|c}{car} & \multicolumn{1}{|c}{person} & \multicolumn{1}{|c}{mIoU ($\%$)}
\\
\midrule
\multicolumn{1}{l|}{DiffuMask} & \multicolumn{1}{c|}{60.0$k$} & \multicolumn{1}{c|}{R50} & \multicolumn{1}{c|}{43.4$\%$} & \multicolumn{1}{c|}{67.3$\%$} & \multicolumn{1}{c|}{60.2$\%$} & \multicolumn{1}{c}{57.0$\%$}\\
\multicolumn{1}{l|}{\textbf{SeeDiff (ours)}} & \multicolumn{1}{c|}{60.0$k$} & \multicolumn{1}{c|}{R50} & \multicolumn{1}{c|}{55.8$\%$} & \multicolumn{1}{c|}{73.6$\%$} & \multicolumn{1}{c|}{68.2$\%$} & \multicolumn{1}{c}{\textbf{65.8$\%$}}\\
\midrule
\multicolumn{1}{l|}{DiffuMask} & \multicolumn{1}{c|}{60.0$k$} & \multicolumn{1}{c|}{Swin-B} & \multicolumn{1}{c|}{72.8$\%$} & \multicolumn{1}{c|}{73.4$\%$} & \multicolumn{1}{c|}{62.6$\%$} & \multicolumn{1}{c}{69.6$\%$}\\
\multicolumn{1}{l|}{\textbf{SeeDiff (ours)}} & \multicolumn{1}{c|}{60.0$k$} & \multicolumn{1}{c|}{Swin-B} & \multicolumn{1}{c|}{83.4$\%$} & \multicolumn{1}{c|}{75.1$\%$} & \multicolumn{1}{c|}{91.0$\%$} & \multicolumn{1}{c}{\textbf{83.2$\%$}}\\

\midrule %
\end{tabular}
}
\caption{\textbf{Semantic segmentation on selected class, ADE-20$k$ }$\texttt{val.}$ }
\label{table:ade20k}
\end{table}

\subsection{Domain Generalization} 
We evaluate the performance for domain generalization at Table~\ref{table:domain_generalization}.
We fix the text input with a basic prompt `a photo of a $<\texttt{CLASS}>$'. 
We evaluate the generated dataset on the validation sets of both Pascal VOC 2012 and Cityscapes, using Mask2Former with a ResNet-50 backbone as the evaluator, following the same settings as our baseline, DiffuMask.

\input{PAMR}

\subsection{Qualitative Results}

\subsubsection{Effect of Seed Region Expansion}
\begin{figure*}[h!]
\centering
\includegraphics[width=1.0\textwidth]{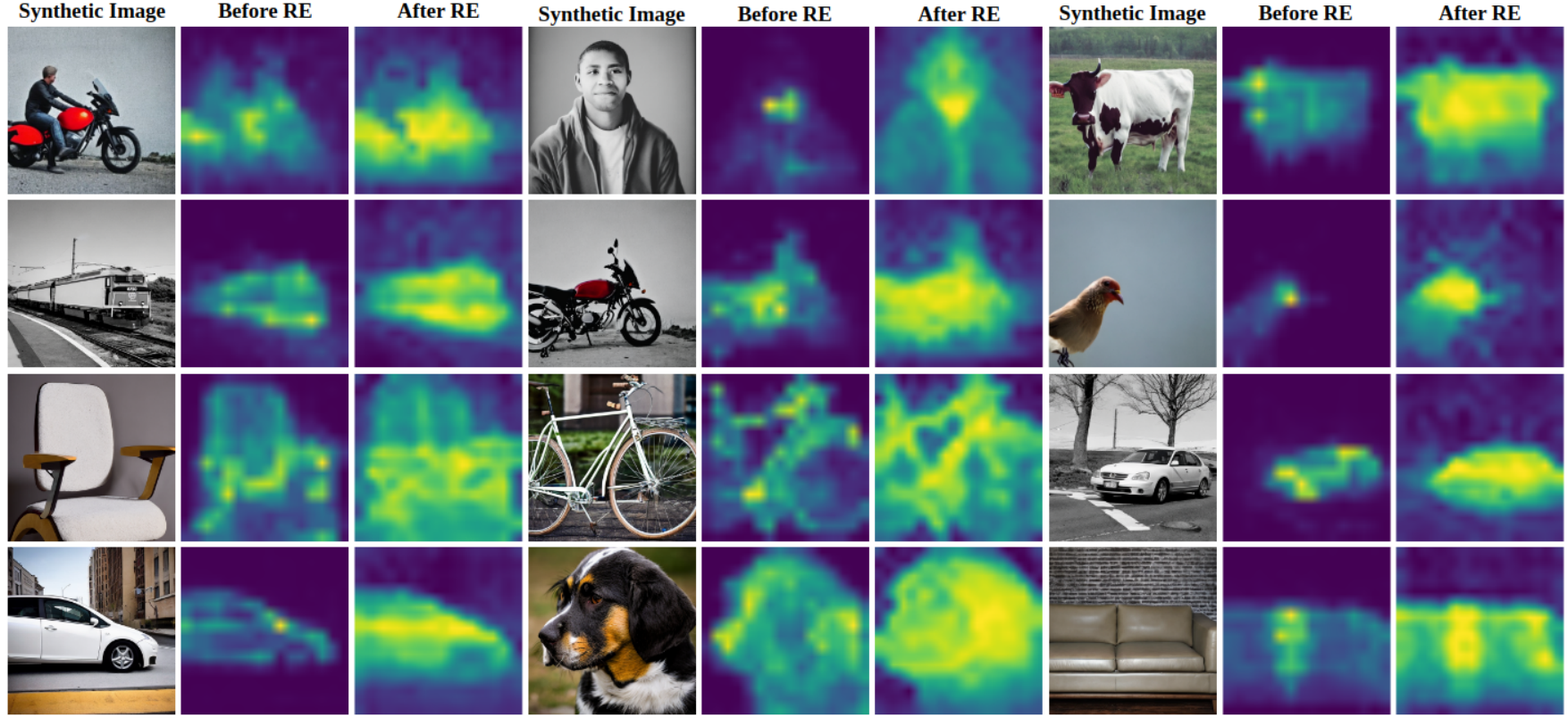}
\caption{\textbf{Effect of seed region expansion process.}}
\label{fig_re}
\end{figure*}

Fig.~\ref{fig_re} presents qualitative results across various images, demonstrating how our method, seeds guided region expansion, compensates for the objectness may missed by the cross-attention aggregation map.

\subsubsection{Effect of Mask Refinement with Background Expansion}
\begin{figure*}[h]
\centering
\includegraphics[width=1.0\textwidth]{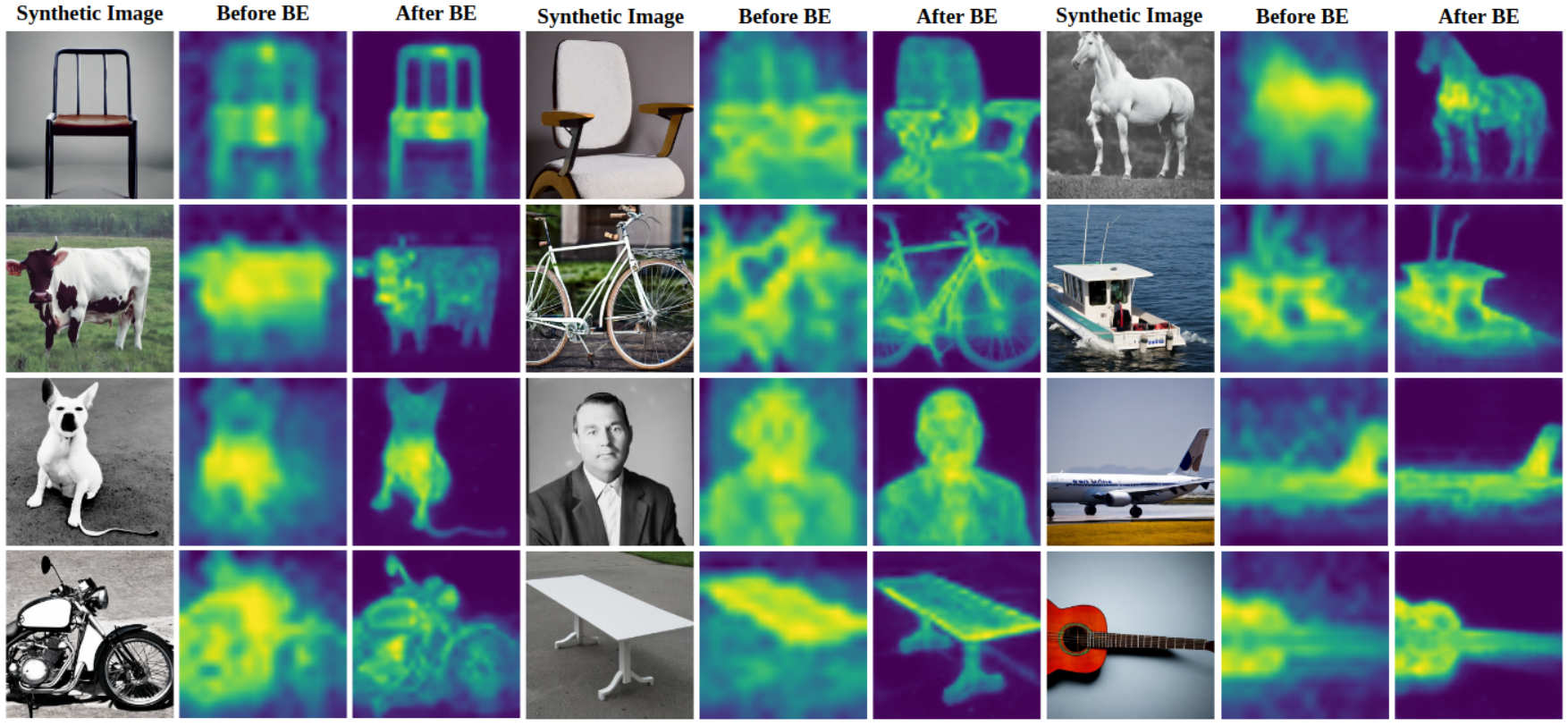}
\caption{\textbf{Effect of refinement with background expansion process.}}
\label{fig_be}
\end{figure*}

Fig.~\ref{fig_be} presents qualitative results across various images, demonstrating how our method, mask refinement with background expansion, that refine the coarse map, using the high resolution of fine-grained self-attention feature map.
Fig.~\ref{fig_be} shows the visualization of the final mask $\mathcal{M}_\text{final}$ before applying the PAMR algorithm, which is part of the mask refinement process. We qualitatively observe that the background expansion effectively refines the coarse mask across images from various classes.
\subsubsection{Comparison with Previous Works : Basic Prompt} 
\begin{figure*}[t]
\centering
\includegraphics[width=1.0\textwidth]{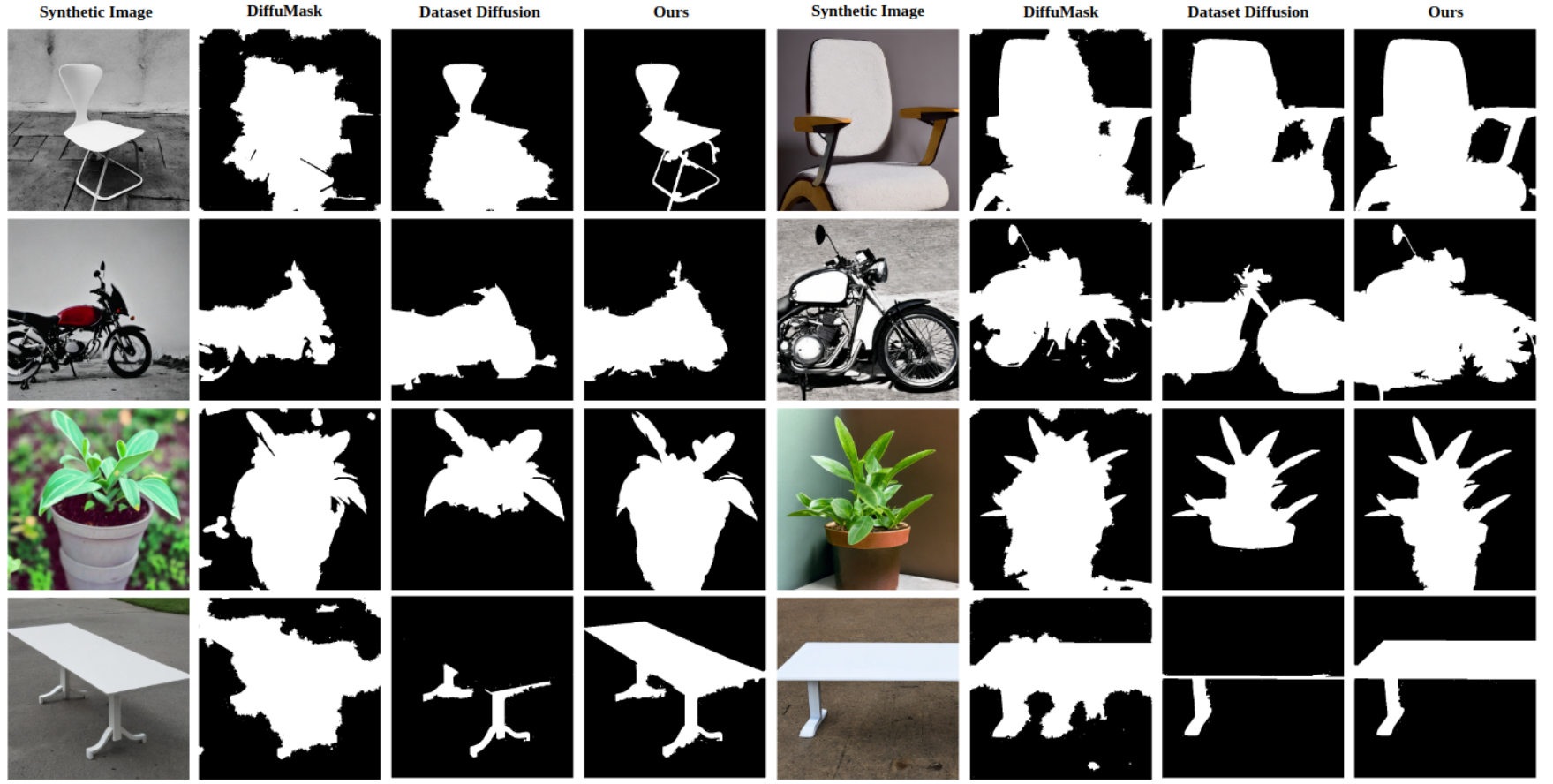}
\caption{\textbf{Refined mask compare with previous works from basic prompt.}}
\label{fig_mask_compare}
\end{figure*}

Fig.~\ref{fig_mask_compare} shows the comparison with previous works, DiffuMask and Dataset Diffusion. 
In the case of Diffumask setting, the final map was derived by aggregating the cross-attention maps across all resolutions. 
For Dataset Diffusion, the final map was generated by first producing the cross-attention aggregation map at a resolution of 16 $\mathcal{A}^{16}_{ca}$ and the self-attention aggregation map $\mathcal{A}^{32}_{sa}$ at a resolution of 32, then multiplying these maps and applying exponentiation using the value of $\tau = 4$.
We applied PAMR consistently in the mask post-processing stage across all methods to compare the masks generated under the same conditions using the basic prompt `a photo of a $<\text{class}>$' without any prompt tuning. 
Additionally, we uniformly set the threshold parameter for generating the binary mask input to PAMR at $\beta=0.3$.

\subsubsection{Open-Vocabulary Mask Generation}
\begin{figure*}[t]
\centering
\includegraphics[width=1.0\textwidth]{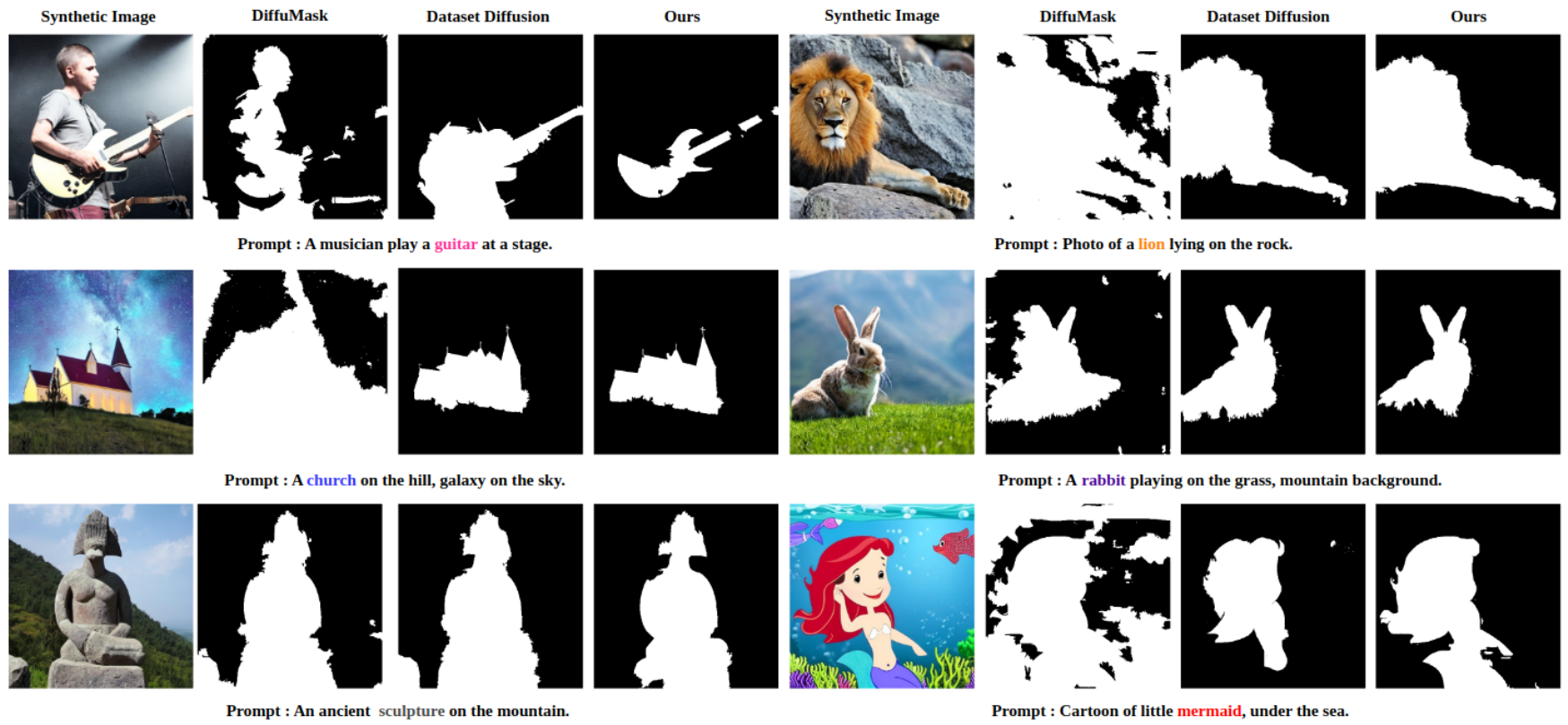}
\caption{\textbf{Qualitative comparisons on masks generated by our SeeDiff against baselines (DiffuMask and Dataset Diffusion), with open-vocabulary prompts.}}
\label{fig_mask_compare_ov}
\end{figure*}

\begin{figure*}[t]
\centering
\includegraphics[width=1.0\textwidth]{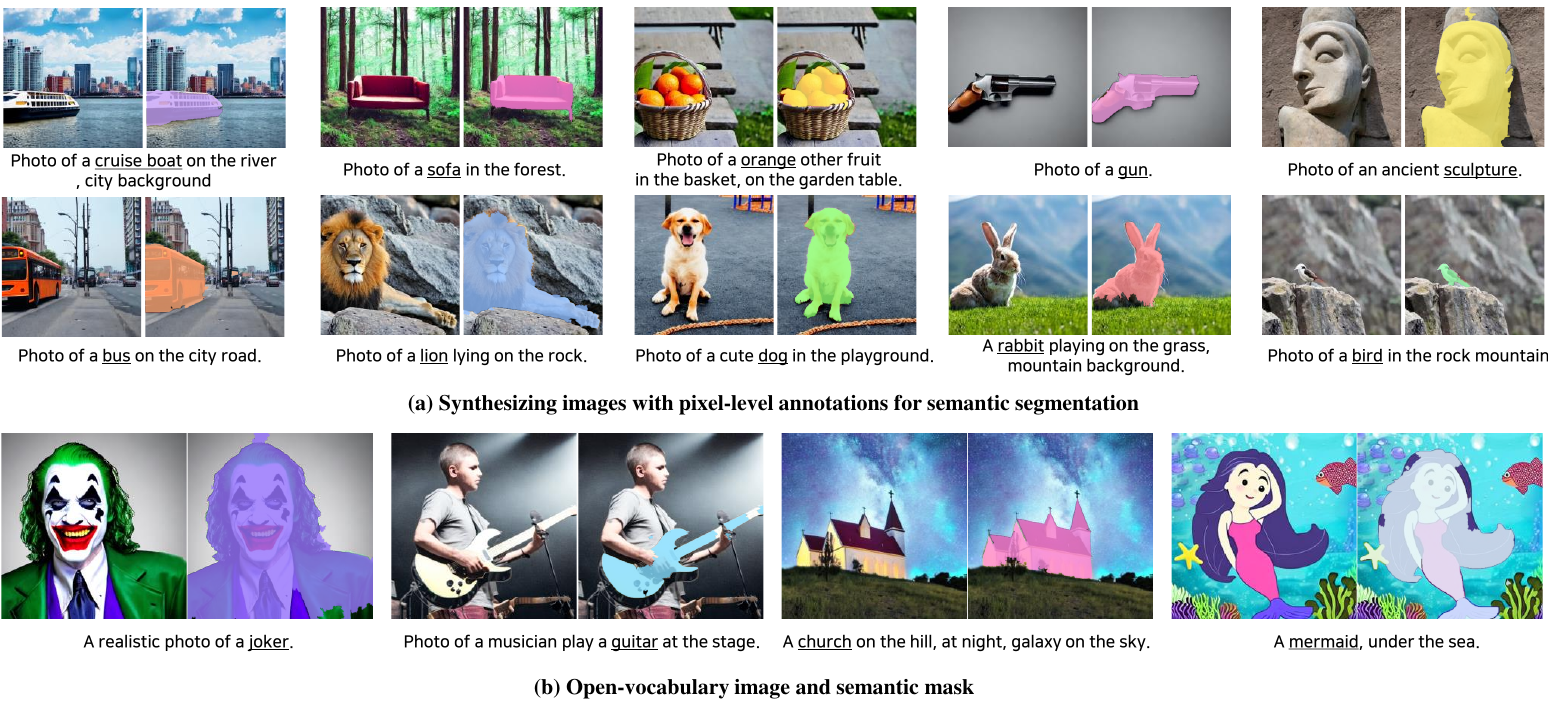}
\caption{\textbf{Qualitative results of masks generated by our method, SeeDiff, with open-vocabulary prompts.}}
\label{fig_mask_compare_ov_2}
\end{figure*}

Not only our methodology but also previous works utilize Stable Diffusion as the backbone model, given its excellent generalization performance. 
This strong generalization capability allows the model to generate masks for the target class even when various sentences with different expressions are used as input text in an open-vocabulary setting.
Therefore, we qualitatively demonstrate the robustness of mask generation for the target class, even with various text inputs in Fig.~\ref{fig_mask_compare_ov}.
And also, in Fig.~\ref{fig_mask_compare_ov_2}, we visualize our generated mask with various open-vocabulary text prompts.

\nocite{*}

\newpage
\clearpage

\input{}

%% file: PAMR.tex
\begin{table}[t]
\centering
\resizebox{.35\textwidth}{!}{ %
\begin{tabular}{@{}lccc@{}}
\toprule
\multicolumn{1}{l}{Method} & \multicolumn{1}{c|}{} & \multicolumn{1}{c|}{PAMR} & \multicolumn{1}{c}{mIoU (\%)} \\   \midrule
\multicolumn{1}{l}{\multirow{2}{*}{DiffuMask}}{}  & \multicolumn{1}{c|}{} & \multicolumn{1}{c|}{\ding{55}} & 66.3\\

\multicolumn{1}{l}{}{}  & \multicolumn{1}{c|}{} & \multicolumn{1}{c|}{\ding{51}} & 68.8\\
\midrule
\multicolumn{1}{l}{\multirow{2}{*}{Dataset Diffusion}}  & \multicolumn{1}{c|}{} & \multicolumn{1}{c|}{\ding{55}} & 69.7 \\
\multicolumn{1}{l}{}  & \multicolumn{1}{c|}{} & \multicolumn{1}{c|}{\ding{51}} & 72.0 \\
\midrule
\multicolumn{1}{l}{\multirow{2}{*}{SeeDiff (Ours)}}  & \multicolumn{1}{c|}{} & \multicolumn{1}{c|}{\ding{55}} & 74.7 \\
\multicolumn{1}{l}{}  & \multicolumn{1}{c|}{} & \multicolumn{1}{c|}{\ding{51}} & 79.6 \\
\midrule
\end{tabular} %
}
\caption{\textbf{Effect of mask refinement post-processing with PAMR on Pascal-VOC 2012} $\texttt{val.}$ }
\label{table:pamr}
\end{table}

\subsection{Mask Post-Processing}
A pre-binarized soft mask (a.k.a. probability map) obtained from attention responses can be coarse and noisy, leading to an uncertainty over how to set a threshold for binarization. 
Both of our baselines, DiffuMask~\cite{diffumask} and Dataset Diffusion~\cite{Dataset-Diffusion}, note that an optimal threshold differs for each class.
To handle such uncertainty, they both perform mask refinement post-processing.
DiffuMask utilizes semantic affinity learning via AffinityNet~\cite{AffinityNet} to find an appropriate threshold for each image.
Meanwhile, Dataset Diffusion proposes to self-train a model with uncertainty-aware loss function with masks that have uncertainty regions defined by two thresholds.
While these methods may have addressed the uncertainty regions, they require extra learning processes with either a pre-trained segmentation network or two hyperparameters.
These kinds of extra learning processes can limit generalization performance to other domains.
On the other hand, we aim to perform \textit{training-free} mask refinement post-processing, such that we do not rely on training procedures with a pre-trained network or an extra hyperparameter (other than a standard threshold for binarization).
To this end, we employ a pixel-adaptive mask refinement (PAMR) introduced by~\cite{pamr}.
Similar to AffinityNet~\cite{AffinityNet} utilized in DiffuMask~\cite{diffumask}, PAMR aims to assign the same class labels to nearby pixels that have similar values.
To do so, PAMR iteratively updates each soft mask pixel $m_{i,j}=\gM[i,j]$ with the convex combination of the pixel values of its neighbors $\gN(i,j)$: 
\begin{equation}
    m_{i,j} \leftarrow \sum_{(i',j')\in\gN(i,j)}k(m_{i,j}, m_{i',j'})\cdot m_{i,j},
\end{equation}
where $k(\cdot,\cdot)$ is a kernel function (i.e., similarity function), which is instantiated as a normalized radial basis function (RBF) in PAMR:
\begin{equation}
    k(m_{i,j}, m_{i',j'})=\frac{\exp(\frac{-\left|m_{i,j}- m_{i',j'}\right|}{\sigma^2_{i,j}})}{\sum_{(i'',j'')\in\gN(i,j)}\exp(\frac{-\left|m_{i,j}- m_{i'',j''}\right|}{\sigma^2_{i,j}})},
\end{equation}
where $\sigma^2_{i,j}$ is the locally computed standard deviation of a pixel value.
In this work, we apply PAMR on a soft mask $\gM_\text{final}$ obtained from Stable Diffusion, before the binarization.
As a result, we obtain an enhanced mask quality, revealed by the performance gain with PAMR as demonstrated in Tab.~\ref{table:pamr}.
One may argue that our final performance is largely due to PAMR.
However, Tab.~\ref{table:pamr} shows that our SeeDiff without PAMR outperforms PAMR-enhanced versions of DiffuMask and Dataset Diffusion.

We also note that our method benefits substantially more from PAMR, compared to other methods.
We believe this is because PAMR acts as pixel-level region expansion, similar to our SA-based region expansion.
Specifically, PAMR finds similar regions at pixel level, akin to how self-attention finds similar regions, however at patch level.
Our SA-based region expansion is restricted to $64\times64$ resolution, due to the design of Stable Diffusion, being unable to perform pixel-level region expansion with SA.
Thus, a pixel-level region expansion with PAMR as our last step of iterative region expansion can be beneficial and thus considered as a integral component to our framework that takes an iterative procedure of finding similar regions with increasing resolutions. 
